\documentstyle[12pt,mkstyle,twoside]{article}
\title{Restricted Causal Inference} 

\date{}
\newcommand{\Bem}[1]{}

\begin{document}

\machetitel

%
%
\section{Introduction}
Causally insufficient structures (models with latent or hidden variables, or 
with confounding etc.) 
of joint probability distributions 
have been subject of intense study not only in 
statistics (e.g.\cite{Joreskog:90}, \cite{Bentler:80}) , but also in various 
AI systems \cite{Cooper:92},\cite{Spirtes:93}. 
 In AI, belief networks, being 
representations of joint probability distribution with an underlying directed 
acyclic graph (dag), \cite{Geiger:90}) structure, are paid special attention 
due to the fact that 
efficient reasoning (uncertainty propagation) methods have been developed for 
them \cite{Pearl:88}, \cite{Shachter:90b}. Algorithms have been therefore 
elaborated to acquire 
the belief network structure from data  \cite{Cooper:92}, \cite{Pearl:88}. As 
artifacts due 
to variable hiding negatively influence the performance of derived belief 
networks, models with latent variables have been studied and several 
algorithms for learning belief network structure under causal insufficiency 
have also been developed \cite{Cooper:92}, \cite{Spirtes:93}. 
Regrettably, some of them are known already to be 
erroneous (e.g. IC algorithm of \cite{Pearl:91} - compare Discussion in 
chapter 6 of  \cite{Spirtes:93}, FCI algorithm of 
\cite{Spirtes:93} - compare \cite{Klopotek:93i}) and other not tractable for 
more than 10 variables (e.g. CI \cite{Spirtes:93} - compare chapter 6 
therein). This paper proposes a new algorithm 
for recovery of belief network structure from data 
handling 
hidden variables. It consists essentially in an extension of the CI algorithm 
of \cite{Spirtes:93} by restricting  the  number  of  conditional 
dependencies 
 checked up to k variables and in an extension of the original CI by 
additional 
steps transforming so called partial including path graph into a belief 
network. Its correctness is demonstrated. 
\section{Basic Definitions}
Please refer to \cite{Spirtes:93} (chapter 6) to recall definitions of 
including path graph, partially oriented including path graph, directed path, 
definite discriminating  path, collider. Refer to \cite{Geiger:90} for 
definitions of active and passive trail, d-separation. 

A partially oriented including path graph $\pi$ contains the following types 
of edges unidirectional: $A->B$, bidirectional $A<->B$, partially oriented 
$Ao->B$ and non-oriented $Ao-oB$, as well as some local constraint 
information $A*-\underline{*B*}-*C$
 meaning that edges between A and B and 
between B and C cannot meet head to head at B. (Subsequently an asterisk (*) 
means any orientation of an edge end: e.g. $A*->B$ means either $A->B$ or 
$Ao->B$ or $A<->B$).

Let us introduce some notions specific for r(k)CIA:
\begin{itemize}
\item[(i)] A is r(k)-separated from B given set S ($card(S)\leq k$) iff A and 
B are conditionally independent given S 
\Bem{- conditional independence means  
$\chi 
 ^2$-test does not deny the thesis of independence of variables A and B given 
S. .}
\item[(ii)] In a partially oriented including path graph $\pi$, 
 a node A is called {\em legally removable} iff there exists no 
 local constraint 
information $B*-\underline{*A*}-*C$ for any nodes B and C and there exists no 
edge of the form $A*->B$ for any node B. 
\end{itemize}
\section{The Algorithm}
{\noindent \bf The Restricted-to-k-Variables Causal Inference Algorithm 
(r(k)CIA):}\\
Input: Empirical joint probability distribution\\
Output: Belief network.
\begin{description}
\item[A)] Form the complete undirected graph Q on the vertex set V.
\item[B)] if A and B are r(k)-separated given any subset S of V, remove the 
edge between 
A and B, and record S in Sepset(A,B) and Sepset(B,A). 
\item[C)] Let F be the graph resulting from step B). Orient each edge 
o-o.  For each 
triple of vertices A,B,C such that the pair A,B and the pair B,C are each 
adjacent in F, but the pair A,C are not adjacent in F, orient \Bem{$(C)$} 
A*-*B*-*C as 
$A*->B<-*C$ if and only if B is not in Sepset(A,C), and orient 
 A*-*B*-*C 
as $A*-\underline{*B*}-*C$ if and only if B is in Sepset(A,C).
\item[D)] Repeat
\begin{description}
\item[(D1) if] there is a directed path from A to B, and an edge  A*-*B, 
orient \Bem{$(D_p)$} A*-*B 
as $A*->B$,
\item[(D2) else if]  B is a collider along $<A,B,C>$ in $\pi$, B is adjacent 
to D, A and C are not adjacent, and there exists 
 local constraint  $A*-\underline{*D*}-*C$, then orient \Bem{$(D_s)$} $B*-*D$ 
as $B<-*D$ ,
  \item[(D3) else if] U is a definite discriminating path between A and B for 
M in $\pi$ and 
P and R are adjacent to M on U, and P-M-R is a triangle, then\\
if M is in Sepset(A,B) then M is marked as non-collider on subpath 
$P*-\underline{*M*}-R$\\
else $P*-*AM*-*R$ is oriented \Bem{$(D_d)$} as $P*->M<-*R$,
\item[(D4) else if] $P*-\underline{>M*}-*R$ then orient \Bem{$(D_c)$} as 
$P*->M->R$.
\item[until] no more edges can be oriented.
\item[E)] Orient every edge $Ao->B$ as $A->B$.
\item[F)] 
\Bem{
 Orient all the edges of type $Ao-oB$ either as $A<-B$ or $A->B$ so 
as not to 
violate  $P*-\underline{*M*}-*R$ constraints as follows: 
}
Copy the  partially oriented including path graph $\pi$ onto $\pi'$. \\
Repeat: \\
 In $\pi'$ identify a legally removable node A. Remove it from  $\pi'$ 
together with every edge $A*-*B$ and every constraint 
 with A involved in it. Whenever an edge $Ao-oB$ is removed from $\pi'$, 
orient 
edge $Ao-oB$ in $\pi$ as $A<-B$. \\
Until no more node is left in $\pi'$.
 \end{description}
\item[End of r(k)CIA]
\end{description}
\section{Differences to Spirtes et al. CI Algorithm}
Steps E) and F) constitute an extension of \Bem{(are not present in)} the 
original CI algorithm of 
\cite{Spirtes:93}, bridging the gap between partial including path graph and 
the belief network. Their properties are subject of  \cite{Klopotek:93g} with 
respect to CI. Their correctness for r(k)CIA may be proven in the same way.

Step B) was modified by substituting the term "d-separation" with 
"r(k)-separation". This means that not all possible subsets S of the set of 
all nodes V (with card(S) up to card(V)-2) are tested on rendering nodes A 
and 
B independent, but only those with cardinality 0,1,2,...,k. If one takes into 
account that higher order conditional independences require larger amounts of 
data to remain stable, superior stability of this step in r(k)CIA becomes 
obvious.
 
Step D2) has been modified in that the term "not d-connected" of CI
was substituted 
by reference to local constraints. In this way results of step B) are 
 exploited more thoroughly and in step D) no more reference is made to 
original 
body of data (which clearly accelerates the algorithm). This modification is 
legitimate since all the other cases covered by the concept of  "not 
d-connected" of CI would have resulted in orientation of $D*->B$ already in 
step C). Hence the generality of step D2) of the original CI algorithm is 
not needed here. 
\section{Properties of the Algorithm}
Obviously, the algorithm r(k)CIA will leave some edges actually not present  
in original data. As demonstrated in \cite{Klopotek:93g}, 
 superfluous edges may lead to incorrect belief network recovery. We shall 
show 
therefore that this is not the case with r(k)CIA. 

In \cite{Klopotek:93g} it has been proven that the original CI extended by 
above-mentioned steps E) and F) will produce a dag compatible with the 
original data. Preliminaries for that result are that given the "real" dag G
with visible variables $V_s$ and hidden ones $V_h$ one can define an 
"intrinsic" dag F in $V_s$ indistinguishable from G with respect to 
dependencies and independences within set $V_s$ such that the modified  CI 
algorithm produces a dag statistically indistinguishable from F. 
(This dag F is called "including path graph" in \cite{Spirtes:93}). 
Below we show 
possibility of defining such a dag F for the r(k)CIA algorithm.

Let us define the r(k)-including path graph: G be a dag with a set of hidden 
variables $V_h$ and of visible variables $V_s$.  A graph  $\pi$ be  a 
r(k)-including path graph for G iff its set of nodes is  $V_s$, and an edge
between A and B from $V_s$ exists in  $\pi$ iff no subset S  of $V_s$ with 
 cardinality not exceeding k  does not d-separate nodes A and B in G. This 
edge 
is out of A iff there exists such a subset S' of $V_s$ with cardinality
 not exceeding k-1 that no trail in G from B to A into A is active with 
respect 
to S'. Otherwise this edge is ingoing into A. It is easily demonstrated that 
 every edge in $\pi$ is either unidirectional or bidirectional (no edge is 
left
unoriented). Because  there exists never a trail outgoing from A and outgoing 
from B which is active with respect to an empty set S ($card(S)=0 \leq k$).
 Furthermore, if there is an edge $A->B$ in $\pi$, then there exists a 
directed 
path from A to B in G. This is easily seen: Let S' be  a subset of $V_s$ 
with cardinality
not exceeding k-1 that no path in G from B to A into A is active with respect 
to S'. (1) Then clearly there must exist a trail in G outgoing out of A 
towards B which 
is active with respect to S' (otherwise edge AB would be absent from $\pi$ as 
S' would d-separate A and B). (2) Let us go along this trail in G as long as  
edges along it passing edges from tail to 
head. In this way we either reach B 
(which 
would complete the proof) or stop at a collider along this trail. This  
collider 
must either be in S' or have a successor in S' (as active trail definition 
requires). Let us continue the journey towards the blocking node in S'. The 
node is either not necessary for S' to block all ingoing trails from B to A 
(in this case we remove it from S'  and start the procedure from the beginning
- that is from point (1)) or it is necessary for that purpose. 
(3) In the latter 
case there is a trail 
between B and A ingoing into A this node is blocking. Let us continue our 
 journey along this trail now in the direction where we pass edges from tail 
to 
head (at least one such direction exists). We continue at point (2). As the 
 graph is a dag and the set S' is finite, the procedure is granted to 
terminate 
on reaching node B. This proves our claim. 

Furthermore, in  $\pi$ if we have two edges 
$A->B<-C$ with A and C not adjacent in $\pi$ then no subset S of $V_s$ 
with cardinality not greater than k containing $B$ such that S 
d-separates A and C in G. Because if such a set S existed then the set  
S-\{B\} with cardinality not greater than  k-1 would have to block in G all
trails from A to B into B or all trails from C to B into B (as this is 
required  by  definition   of   d-separation).   But   then   the 
aforementioned 
definition of $\pi$ would require that either edge BA or BC resp. would be 
 out of B.  

Also in  $\pi$  if A,B are adjacent, B,C are adjacent, 
but A,C are not adjacent in $\pi$ and on the trail A-B-C 
node B is non-collider then there exists no such subset S of $V_s$ with 
cardinality not greater than k not containing $B$ that S d-separates A and C 
in G. Otherwise if such a set S existed then (without restriction of 
generality let us assume  $A<-B$) there  exists a directed path from A to B 
in G. The set S would either block it or not. If not, then S would have to 
 block all the trails from C to B which is a contradiction because then edge 
BC 
could not exist in $\pi$. Hence it must block it. But then S would have to 
block every trail ingoing into B either from direction of A or  of C.  
 Should it block those from direction of A (C) then there would exist an 
active 
trail outgoing from B towards A (C) and an active trail between B and C (A). 
But this is a contradiction as then there would exist an active trail 
connecting A and C (via B). This proves our claim. 

Last not least it may be demonstrated that if there exists in 
 $\pi$ a bidirectional edge between A and B, and if there existed an oriented 
path from A to B and if there exists edge  $C*->A$ in
 $\pi$, then in $\pi$ there exists also the edge $C*->B$. This means that a
bidirectional edge $A<->B$ in  $\pi$ can be treated as a 
unidirectional edges   $A<-H->B$, with  H being a parentless hidden variable
and A and B being not adjacent in the graph.
 
The above-mentioned statements indicate that for a faithful graph G 
for edge pair A-B and B-C with A and C not adjacent 
a statistical test of independence of A and C relatively to sets S containing 
B with cardinality not greater than k will correctly decide about orientation 
of edges with respect to  the r(k)-including path graph  $\pi$.
 
In this sense we can prove for $\pi$ that it is a directed acyclic graph. 
 Because graph G is a dag, 
 bidirectional edges in       
  $\pi$ have no impact on ordering, and unidirectional edges in $\pi$ 
must  have orientation in accordance with G as existence of an oriented 
path in G between nodes at both ends of an edge of $\pi$ is assumed. 
With these prerequisites correctness proof of a modified version of CI 
algorithm 
from \cite{Klopotek:93g} extends straight forward to r(k)CIA algorithm
\section{Discussion and Concluding Remarks}
Within this paper a new algorithm of recovery of belief network structure 
from data has been presented and its correctness demonstrated. It relies 
essentially on "acceleration"  of the known CI 
algorithm of Spirtes, Glymour and Scheines \cite{Spirtes:93}
 by restricting the number of conditional dependencies 
checked up to k variables
and it extends CI by additional 
steps transforming so called partial including path graph into a belief 
network. Sample outputs of CI, r(1)CIA and r(2)CIA are shown in Fig.1. 
Though r(k)CIA introduces redundant edges (e.g. AC and FC in Fig.1b), 
indicating dependencies not present in 
the original data, it actually avoids pitfalls of the  FCI algorithm, another 
CI "accelerator" proposed by   Spirtes, Glymour and Scheines 
\cite{Spirtes:93}, as visible from section 5 and \cite{Klopotek:93g}.

\input TO_PIC.INI
\begin{figure}
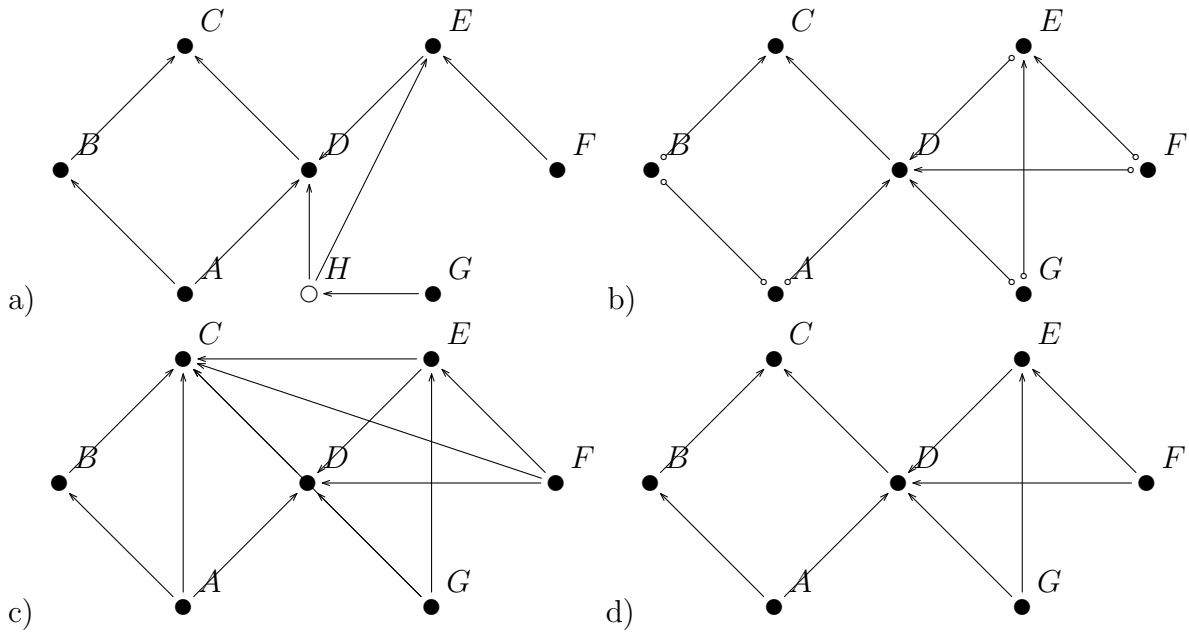

\input RKCI.PIC
\caption{a) Original dag, b) CI output, c) r(1)CIA output, d) r(2)CIA output}
\end{figure}

\newcommand{\LitStelle}[2]{  

\vspace{-2.5mm}

\bibitem{#1}  }
%


 \end{document}